# The Rise of Diffusion Models in Time-Series Forecasting

Caspar Meijer
EEMCS Distributed Systems
Delft University of Technology
`c.j.meijer-1@student.tudelft.nl`

Lydia Y. Chen
EEMCS Distributed Systems
Delft University of Technology
`lydiaychen@ieee.org`

15 Januari 2023

**Abstract**

This survey delves into the application of diffusion models in time-series forecasting. Diffusion models are demonstrating state-of-the-art results in various fields of generative AI. The paper includes comprehensive background information on diffusion models, detailing their conditioning methods and reviewing their use in time-series forecasting. The analysis covers 11 specific time-series implementations, the intuition and theory behind them, the effectiveness on different datasets, and a comparison among each other. Key contributions of this work are the thorough exploration of diffusion models' applications in time-series forecasting and a chronologically ordered overview of these models. Additionally, the paper offers an insightful discussion on the current state-of-the-art in this domain and outlines potential future research directions. This serves as a valuable resource for researchers in AI and time-series analysis, offering a clear view of the latest advancements and future potential of diffusion models.

## 1 Introduction

The advent of generative artificial intelligence (AI) has been transformative across various domains, ranging from education [2, 3, 4] to workplaces [5, 6] and daily activities [7]. Central to this transformation is deep learning, a key pillar enabling AI to analyze and synthesize complex data patterns. Initially, generative AI is defined by its ability to create new, original data samples that reflect the statistical characteristics of a specified dataset, represented mathematically as: given a sample $x$ from distribution $q(x)$, the generative model produces outputs $\hat{x}$ that appear to be drawn from $q(x)$ [8].

Following this introductory note on generative AI, the focus shifts to time-series forecasting, an area of critical importance across various industries. Time-series data, characterized by their temporal dependencies and multifaceted interactions, present unique challenges and opportunities for forecasting future events based on historical data. This is particularly significant in fields such as healthcare prediction [9, 10, 11, 12, 13, 14], energy management [15, 16, 17], and traffic control [18, 19].

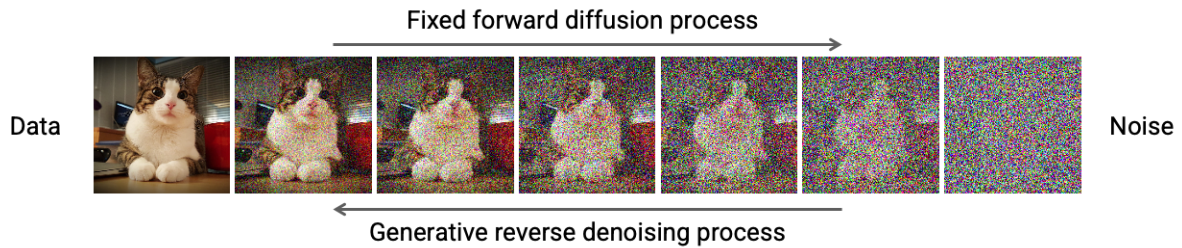

Figure 1: Denoising Diffusion Process [1]

In the landscape of time-series forecasting, the solution space has evolved considerably over time. Initial advancements were marked by the introduction of Long Short-Term Memory (LSTM) variants, notably, the Seq2Seq Autoencoder-LSTM [20]. However, a significant paradigm shift occurred in 2017 with the introduction of the Transformer structure, which incorporated attention mechanisms [21]. This innovation addressed the critical limitation of LSTMs, which was the loss of previous information over extended sequences [22]. Subsequent developments in Transformer-based models, [23, 24, 25, 26, 27, 28], have furthered the field.

In the field of generative modeling, the introduction of modeling structures such as variational autoencoders (VAEs), normalizing flows (NFs), and generative adversarial networks (GANs) have marked significant advancements [29, 30]. Yet, the emergence of diffusion models signals a revolutionary period, promising superior-quality outputs that are pushing the state-of-the-art [31, 32, 33]. A diffusion model can be characterized by the fact that they simulate, as the name implies, a diffusion process transforming data into white noise and then reversing it back into data as shown in Figure 1. These models, capable of approximating the original data distribution, have shown exceeding results in various domains, including image [34, 31, 35, 36], text [37, 38, 39], speech [40, 41] and video synthesis [42, 43, 44, 45].

In this paper, we narrow our focus to the application of diffusion models in time-series forecasting. These models, distinguished by their profound ability to comprehend intricate data dynamics, are revolutionizing this field [46, 47]. First time-series forecasting is further formalized together with how to evaluate the models in sub-section 1.1. Afterwards, the intrinsic workings of the diffusion model and how to condition it is explained in section 2. Then the diffusion-based time-series forecasting papers are discussed in section 3. Finally, a comprehensive discussion and future works is given in sections 4 and 5 respectively.

The main contributions of this paper are:

- In-depth preliminary section about diffusion models and the methods of conditioning that are used for time-series modeling.

- A chronologically ordered overview of the diffusion models capable of time-series forecasting, going in-depth on the implementation in relation to the preliminary, specify the results on which datasets, and a discussion about the work in relation to other diffusion models.

- The work serves as in depth overview for researchers seeking to acquaint themselves with the methods that make up the current state-of-the-art diffusion models for time-series forecasting.

## 1.1 Time-series

Time-series modeling, as highlighted by Koo and Kim [47] and Lin *et al.* [46], is a specialized form of conditional generative modeling, where segments of the time-series are used to generate other segments. This area encompasses three key types: generation, imputation, and forecasting. Generation is about creating synthetic time-series data; imputation deals with filling gaps in existing data, and forecasting is the prediction of future values. These types are interconnected, with forecasting being a specific form of imputation and both imputation and forecasting being aspects of generation. This section will delve into the problem definition for time-series forecasting and evaluation metrics to assess model performance.

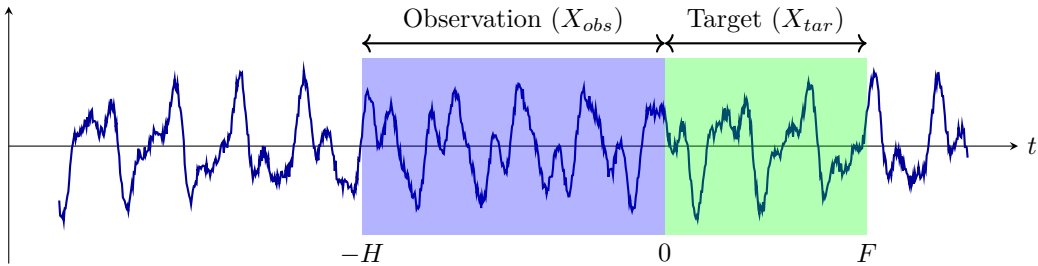

Figure 2: Example of univariate time-series forecasting situation.

### 1.1.1 Problem definition for Time-series Forecasting

Since time is inherently continuous, these series are represented through discrete time steps, determined by a chosen sampling frequency, which can range from milliseconds to hours. For simplicity we assume that the entire dataset $X$ is discretized evenly in the rest of the paper if not explicitly specified. The historical time range, comprising $H$ steps, is defined as the set of observations $obs = \{t \in \mathbb{Z} \mid -H < t \leq 0\}$, while the future range, spanning $F$ steps, is denoted as the targets $tar = \{t \in \mathbb{Z} \mid 0 < t \leq F\}$. Consequently, historical data can be represented as $X_{obs} \in \mathbb{R}^{d \times H}$, and the target forecast data as $X_{tar} \in \mathbb{R}^{d \times F}$, where $d$ indicates the number of distinct features. A time-series is classified as univariate when $d = 1$ as illustrated in Figure 2, and as multivariate when $d > 1$. In this context, a specific value at time step $t$ and feature $i$ from the time-series can be defined as $x_{i,t}$, where $i \in \{1 \ldots d\}$ and $t \in \{-H, \ldots, F\}$ or as $x_t$ when talking about the entire value vector at the time step.

### 1.1.2 Evaluation Metrics

The evaluation of time-series forecasting can be done through a varaiaty of metrics. The most common ones are through the use of the Mean Squared Error (MSE), Mean Absolute Error (MAE), and the Continuous Ranked Probability Score (CRPS) [48]. The biggest difference between these metrics is that the MSE & MAE focus on the mean error, while the CRPS also takes into account the uncertainty of the prediction.

The MSE and MAE over a time-series forecasting are defined as:

$$\text{MSE}(\hat{X}_{tar}, X_{tar}) = \frac{1}{F} \sum_{t=0}^{F} (x_t - \hat{x}_t)^2 \qquad \text{MAE}(\hat{X}_{tar}, X_{tar}) = \frac{1}{F} \sum_{t=0}^{F} |x_t - \hat{x}_t| \qquad (1)$$

The output of these error measurements are vectors for the multivariate time-series, at end the average over all features is taken as the final value. One must consider that the values ought to be normalized, as otherwise errors of different features can have very different scales.

The CRPS is a metric that is used for probabilistic forecasting and measures the compatibility of a cumulative distribution function (CDF) $F$ with an observation $X_{tar}$. In other words, the CRPS metric becomes smaller if the distribution is highly concentrated on the prediction as illustrated in Figure 3. However, in many scenarios the CDF $F$ is not available analytically, but only by estimating it through a set of $N$ forecast samples $\hat{X}_{tar}^N$ [49]. These samples are gathered by sampling the probabilistic model $N$ times. The CRPS metric is calculated for of a single feature at a single timestamp. It is possible to estimate the empirical CDF $\hat{F}_N(z)$ given a point $z$ from these predictions. The empirical CDF at a point $z$ is estimated by the proportion of forecast values that are less than or equal to $z$ and is defined as

$$\hat{F}_{i,t}^N(z) = \frac{1}{N} \sum_{n=1}^{N} \mathbb{I}\{\hat{x}_{i,t}^n \leq z\}. \qquad (2)$$

Where $\mathbb{I}\{\hat{X}_{i,t}^i \leq z\}$ is the indicator function that equals 1 if the condition $\hat{X}_{tar}^i \leq z$ is true, and 0 otherwise. The CRPS for the empirical CDF $\hat{F}_N$ and the observation $X_{tar}$ is then calculated as:

$$\text{CRPS}(\hat{F}_{i,t}^N, x_{i,t}) = \int_{\mathbb{R}} \left( \hat{F}_{i,t}^N(z) - \mathbb{I}\{x_{i,t} \leq z\} \right)^2 dz \qquad (3)$$

Where $i$ indicates the feature and $t$ the timestamp of the forecasts. The estimation of this formulation is further explained and optimized by Jordan *et al.* [49].

As previously mentioned, the CRPS focuses on a single feature at a specific timestamp. To transform this for a multivariate time-series forecast with multiple timestamps in the future some normalization and averaging ought to be done. This can be either done as the Normalized Average CRPS, which is formulated as such:

$$\text{NACRPS}(\hat{F}^N, X_{tar}) = \frac{\sum_{i,t} \text{CRPS}(\hat{F}_{i,t}^N, x_{i,t})}{\sum_{i,t} |x_{i,t}|} \qquad (4)$$

And then there is the CRPS-sum, which is the CRPS for the distribution F for the sum of all $d$ features:

$$\text{CRPS}_{\text{sum}}(\hat{F}^N, X_{tar}) = \frac{\sum_t \text{CRPS}(\hat{F}_{i,t}^N, \sum_i x_i)}{\sum_{i,t} |x_{i,t}|} \qquad (5)$$

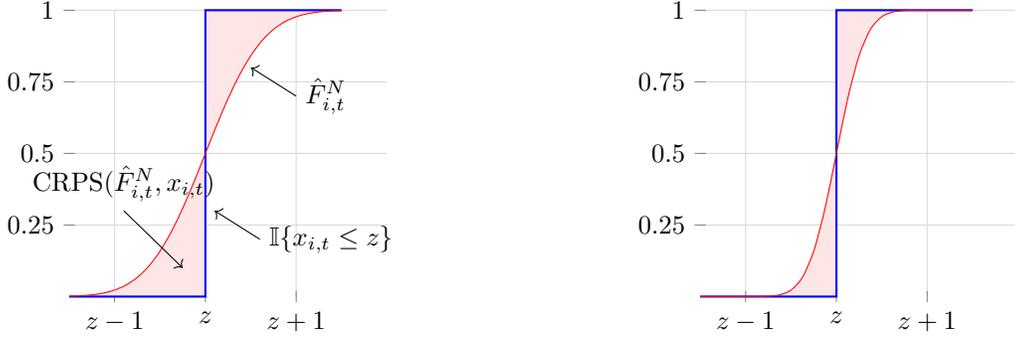

Figure 3: Comparative Visualization of Continuous Ranked Probability Scores: The left plot exhibits a higher CRPS value, indicated by the greater area between the forecast distribution (red line) and the step function (blue line), compared to the right plot.

## 2 Preliminary: Diffusion Models

The foundational works of diffusion models were established by Ho *et al.* [50] and Song *et al.* [51]. These foundations are the unconditional generation of data, which means to create data samples without relying on specific conditions, such as text prompts. Formally, unconditional generation can be described as viewing the training data $x$ as a distribution $q(x)$ from which it is possible to extract samples described by $x \in \mathcal{R}^d$. Now the goal is to approximate this distribution as $p_\theta(x)$ and be able to sample new unseen data from this approximation [8].

Diffusion models approximate the distribution by learning how data can be recovered after it has been diffused to pure noise. The model attempts to transform a Gaussian distribution back into the data distribution as illustrated in Figure 4. This process enables the model to generate data samples from noise, noise is turned into data that resembles the training dataset.

Where Ho *et al.* [50] describes the diffusion and denoising processes as discrete steps and Song *et al.* [51] generalizes these processes to continuous time with the use of stochastic differential equations (SDE). The discrete implementation is described in section 2.1 and the continuous version in section 2.2. However, unconditional generation is not useful if specific data samples are desired. For instance, generating images that contain specifics that are described through text prompts [35]. Such conditional generation of data is further explained in section 2.3

### 2.1 Denoising Diffusion Probabilistic Model (DDPM) (2020) [50]

Core to these models are two pivotal processes: the forward diffusion and the reverse denoising process.

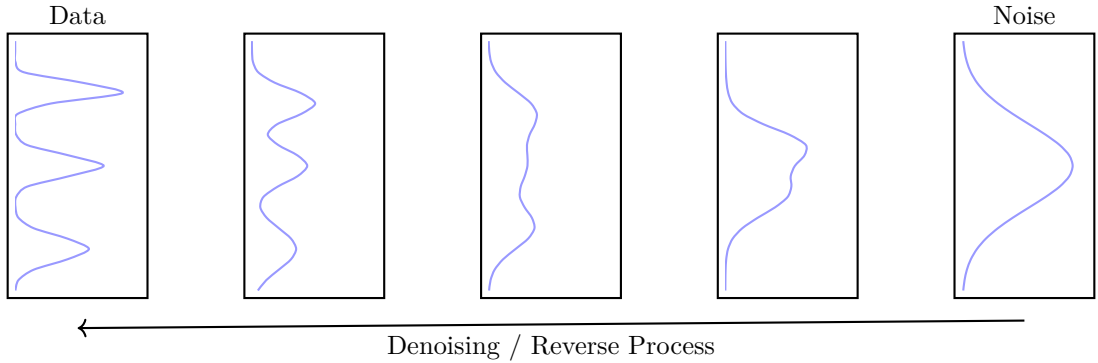

Figure 4: Probability Distribution Evolution: From Noise to Data

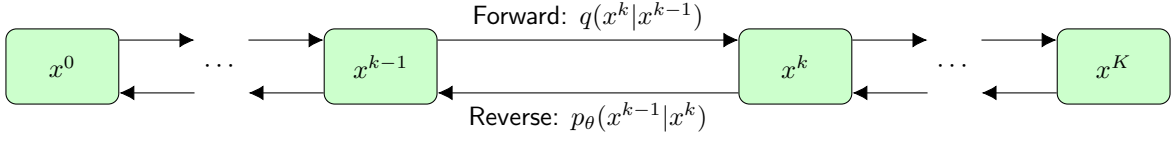

Figure 5: Illustration of the Denoising Diffusion Probabilistic Model process

The forward process transforms an input $x^0$ into a noise vector $x^K$ through a Markov Chain, progressively introducing noise over $K$ steps as can be seen in Figure 5. The transformation is described as:

$$q(x^{1:K}|x^0) = \prod_{k=1}^{K} q(x^k \mid x^{k-1}) \text{ where } q(x^k|x^{k-1}) = \mathcal{N}(x^k; \sqrt{1-\beta_k}x^{k-1}, \beta_k \mathbf{I}) \tag{6}$$

where $\beta_k \in [0,1]$ dictates the variance of the noise introduced at each stage. Obtaining $x^k$ is achieved through:

$$q(x^k|x^0) = \mathcal{N}(x^k; \sqrt{\alpha_k}x^0, (1-\alpha_k)\mathbf{I}) \tag{7}$$

where $\alpha_k = \prod_{i=1}^{k}(1-\beta_i)$ and can be also represented in a closed form as:

$$x^k(x^0, \epsilon) = \sqrt{\alpha_k}x^0 + \sqrt{1-\alpha_k}\epsilon \quad \text{where } \epsilon \sim \mathcal{N}(0, \mathbf{I}) \tag{8}$$

The reverse process transforms the noised data $x^k$ back to its input $x^0$. It is defined by the following Markov chain with $x^K \sim \mathcal{N}(0, \mathbf{I})$:

$$p_\theta(x^{0:K}) = p(x^K) \prod_{k=1}^{K} p_\theta(x^{k-1} \mid x^k) \text{ with } p_\theta(x^{k-1}|x^k) = \mathcal{N}(x^{k-1}; \mu_\theta(x^k, k), \sigma_k^2 \mathbf{I}) \tag{9}$$

In this backwards process $\mu_\theta(x^k, k)$ is modeled using a neural network and where

$$\sigma_k^2 = \begin{cases} \beta_k & \text{optimal for } x^0 \sim N(0, \mathbf{I}) \\ \tilde{\beta}_k = \frac{1-\alpha_{k-1}}{1-\alpha_k}\beta_k & \text{optimal for } x^0 \text{ deterministically set to one point.} \end{cases} \tag{10}$$

The choices of $\sigma_k^2$ correspond to the upper and lower bounds on reverse process entropy for data with coordinatewise unit variance [52].

Training the diffusion model involves minimizing a KL-divergence loss:

$$\mathcal{L}_k = D_{KL}\left(q(x^{k-1}|x^k)||p_\theta(x^{k-1}|x^k)\right) \tag{11}$$

For more stability in training and since $x^0$ is known, the forward process $q(x^{k-1}|x^k)$ is replaced by:

$$q(x^{k-1}|x^k, x^0) = \mathcal{N}(x^{k-1}; \tilde{\mu}_k(x^k, x^0), \tilde{\beta}_k(x^k, k)) \tag{12}$$

$$\text{where } \tilde{\mu}_k(x^k, x^0) = \frac{\sqrt{\alpha_{k-1}}\beta_k}{1-\alpha_t}x^0 + \frac{\sqrt{(1-\beta_k)}(1-\alpha_{k-1})}{1-\alpha_k}x^k \tag{13}$$

$$\text{and } \tilde{\beta}_k = \frac{1-\alpha_{k-1}}{1-\alpha_k}\beta_k \tag{14}$$

This modification allows to reparameterize the training objective from Equation 11 to:

$$\mathcal{L}_k = \frac{1}{2\sigma_k^2}||\tilde{\mu}_k(x^k, x^0) - \mu_\theta(x^k, k)||^2 \tag{15}$$

Where $x^k$ is obtained through Equation 8. From here, $\mu_\theta(x^k, k)$ can be expressed either by a data prediction model $x_\theta(x^k, k)$ or a noise prediction model $\epsilon_\theta(x^k, k)$ [50, 53, 54]. For image generation, Ho *et al.* [50] found that the latter performs better. Using a noise prediction model with noise loss function:

$$\mu_\epsilon(\epsilon_\theta) = \frac{1}{\sqrt{1-\beta_k}}\left(x^k - \frac{\beta_k}{\sqrt{1-\alpha_k}}\epsilon_\theta(x^k, k)\right) \text{ with } \mathcal{L}_{\epsilon_\theta} = \mathbb{E}_{k,x^0,\epsilon}\left[||\epsilon - \epsilon_\theta(x^k, k)||^2\right] \tag{16}$$

Alternatively, using the data prediction model and its data loss function:

$$\mu_x(x_\theta) = \frac{\sqrt{1-\beta_k}(1-\alpha_{k-1})}{1-\alpha_k}x^k + \frac{\sqrt{\alpha_{k-1}}\beta_k}{1-\alpha_k}x_\theta(x^k,k) \text{ with } \mathcal{L}_{x_\theta} = \mathbb{E}_{k,x^0,\epsilon}\left[||x^0 - x_\theta(x^k,k)||^2\right] \quad (17)$$

For an elaborate derivation of all of the equations mentioned above, take a look at the paper by Luo [8]. The actual sampling of $x^{k-1}$ is shown in Figure 6 where $\sigma_k$ is defined as in Equation 10 and $\mathbf{z} \sim \mathcal{N}(0,\mathbf{I})$ Furthermore, Benny and Wolf [53] suggest that the combination of the noise and data methods can enhance performance.

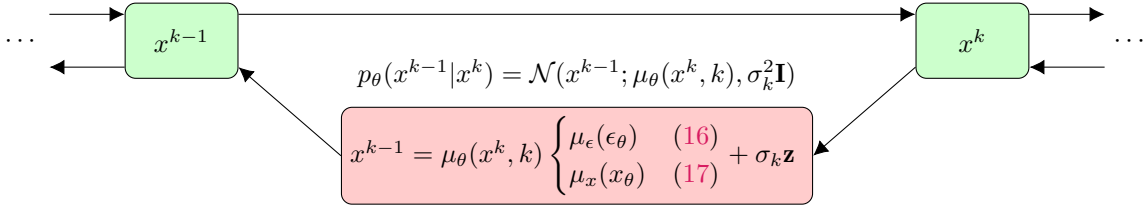

Figure 6: Illustration for prediction models of the reverse process.

## 2.2 Score-based generative modeling through SDE (2021) [51]

DDPMs operate in a discrete space, with discrete diffusion and denoising steps. With the use of stochastic differential equations (SDE), this process can be described in continuous time as shown in Figure 7. This can be thought of as a generalization of the DDPM, as that is a discrete form of SDE [51]. Keep in mind that even though the score-based generative modeling with SDEs is described in continuous time, the solutions are still solved iteratively [51].

The process can be formalized by letting $\mathbf{w}$ and $\bar{\mathbf{w}}$ be a standard Wiener process and its time-reverse version, respectively, and consider a continuous diffusion time $k \in [0,K]$. The forward diffusion process is described with an SDE as:

$$\mathrm{d}x = f(x,k)\mathrm{d}k + g(k)\mathrm{d}\mathbf{w} \quad (18)$$

where $g(k)\mathrm{d}\mathbf{w}$ is the stochastic diffusion and $f(x,k)\mathrm{d}k$ is the deterministic drift. On the other hand, the reverse denoising process is described with a time-reverse SDE [55]:

$$\mathrm{d}x = \left[f(x,k) - g(k)^2 \nabla_x \log p_k(x)\right]\mathrm{d}k + g(k)d\bar{\mathbf{w}} \quad (19)$$

where the score of the marginal distribution $\nabla_x \log p_k(x)$ is to be estimated. It is estimated with a time-dependent score-based model $s_\theta(x,k)$ using the score objective function

$$\mathcal{L}_{s_\theta} = \mathbb{E}_{k,x(0),x(k)}\left[||\nabla_{x(k)} \log p_{0k}(x(k)|x(0)) - s_\theta(x(k),k)||^2\right] \quad (20)$$

Using this objective function the score-based model $s_\theta(x,k)$ can be trained on samples with score matching [51, 56, 57, 58, 59].

The above formula is a general representation of a SDE, and now to be more specificly tailored to the diffusion process there are multiple methods as described in Song *et al.* [51], of which 2 are described as

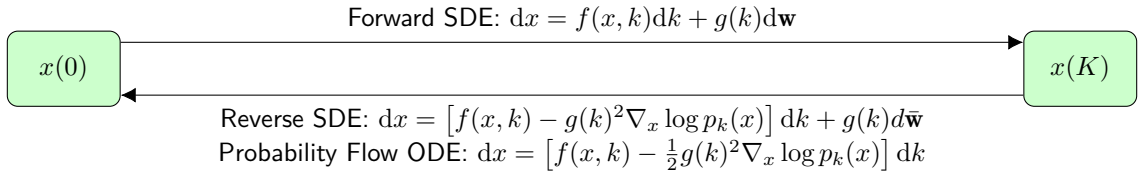

Figure 7: Illustration for prediction models of the reverse process.

following. The DDPM process can also be generalized to a SDE and is termed the variance preserving (VP) SDE:

$$\mathrm{d}x = -\frac{1}{2}\beta(k)x\mathrm{d}k + \sqrt{\beta(k)}\mathrm{d}\mathbf{w} \tag{21}$$

where $\beta(\cdot)$ is a continuous function and $\beta(\frac{k}{K}) = K(1-\beta_k)$ from (6) as $K \to \infty$. Based on the VP SDE, Song et al. [51] has designed the sub-VP SDE, specialized for likelihoods:

$$\mathrm{d}x = -\frac{1}{2}\beta(k)x\mathrm{d}k + \sqrt{\beta(k)\left(1 - \mathrm{e}^{-2\int_0^k \beta(s)ds}\right)}\mathrm{d}\mathbf{w} \tag{22}$$

Song et al. [51] demonstrated that from each SDE, one can derive an ordinary differential equation (ODE) with the same marginal distributions. This associated ODE of an SDE is called the *probability flow* ODE. Applying this concept to the reverse-SDE (19) gives the subsequent probability flow ODE:

$$\mathrm{d}x = \left[f(x,k) - \frac{1}{2}g(k)^2\nabla_x \log p_k(x)\right]\mathrm{d}k \tag{23}$$

When the $\nabla_x \log p_k(x)$ in the probability flow is replaced by its approximation $s_\theta(x,k)$, it becomes a special case of neural ODE [60]. Specifically, it resembles continuous normalizing flows because it transforms data distributions to prior noise distributions and it is fully invertible [61]. Furthermore, this allows exact log-likelihood computation and can be trained via maximum likelihood using standard methods [60].

## 2.3 Conditional Diffusion Models

In this section, the details are explained how the diffusion models can be conditioned. Conditioning a generative model means to control the output given some extra input. This input can be text describing an image [35], or part on an image for inpainting [35] The extra input **c** is sometimes called the covariates but often times the conditional input. Diffusion models can be conditioned in two different ways. Firstly in section 2.3.1, during training, by inputting the condition **c** into the model and training accordingly [34]. Secondly in section 2.3.2, there is Diffusion Guidance, a novel approach that shifts the conditioning from training to the inference stage, leveraging the Bayes rule [51].

### 2.3.1 Conditional Denoising Model

Conditional versions of the model can be extended to embrace additional input $c$, altering the backward denoising step in Equation 9 [34]:

$$p_\theta(x^{0:K}|\mathbf{c}) = p(x^K)\prod_{k=1}^{K} p_\theta(x^{k-1} \mid x^k, \mathbf{c}) \text{ with } p_\theta(x^{k-1}|x^k, \mathbf{c}) = \mathcal{N}(x^{k-1}; \mu_\theta(x^k, k|\mathbf{c}), \sigma_k^2\mathbf{I}) \tag{24}$$

Consequently, the noise and data variants of the prediction models are conditioned as respectively:

$$\mu_\epsilon(\epsilon_\theta, \mathbf{c}) = \frac{1}{\sqrt{1-\beta_k}}\left(x^k - \frac{\beta_k}{\sqrt{1-\alpha_k}}\epsilon_\theta(x^k, k|\mathbf{c})\right) \tag{25}$$

$$\mu_x(x_\theta, \mathbf{c}) = \frac{\sqrt{1-\beta_k}(1-\alpha_{k-1})}{1-\alpha_k}x^k + \frac{\sqrt{\alpha_{k-1}}\beta_k}{1-\alpha_k}x_\theta(x_k, k|\mathbf{c}) \tag{26}$$

where the conditioning in practice are extra inputs to the prediction model as illustrated in Figure 8. The same condition is included in each backwards step.

For the score-based generative models through SDE such conditioning also exists, where the conditioning happens in the marginal distributions of the score [62]. The reverse-time SDE becomes:

$$dx = \left[f(x,k) = g(k)^2\nabla_x \log p_k(x|\mathbf{c})\right]dk + g(k)d\bar{w} \tag{27}$$

and the objective function then becomes

$$\mathcal{L}_k = \mathbb{E}_{k,x(0),x(k)}\left[\left|\left|\nabla_{x(k)}\log p_{0k}(x(k)|x(0)) - s_\theta(x(k),k|\mathbf{c})\right|\right|^2\right] \tag{28}$$

.

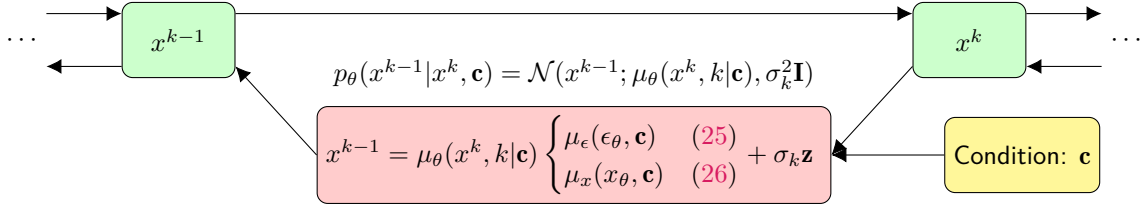

Figure 8: Illustration for conditional prediction models of the reverse process.

### 2.3.2 Diffusion Guidance

In traditional diffusion models, conditioning is typically introduced during the training process, as represented in equation (24). However, a novel approach introduces conditions during the inference stage by employing the concept of guidance [52, 51]. This method is grounded in the application of the Bayes rule to the probability $p(x^k|\mathbf{c})$ given by:

$$p(x^k|\mathbf{c}) = \frac{p(\mathbf{c}|x^k) \times p(x^k)}{p(\mathbf{c})} \tag{29}$$

Taking the natural logarithm of both sides, we can further derive:

$$\log p(x^k|\mathbf{c}) = \log p(\mathbf{c}|x^k) + \log p(x^k) - \log p(\mathbf{c}) \tag{30}$$

From which, upon differentiation with respect to $x^k$, we obtain:

$$\nabla_{x^k} \log p(x^k|\mathbf{c}) = \nabla_{x^k} \log p(\mathbf{c}|x^k) + \nabla_{x^k} \log p(x^k) \tag{31}$$

The relation in (31) provides a bridge to guide the reverse diffusion process in equation (9) and (19). Instead of merely reversing the diffusion, this guides the samples to conform to the desired condition $\mathbf{c}$ [52, 51]. Thus, equation (9) can be conditioned as:

$$p_\theta(x^{k-1}|x^k, \mathbf{c}) = \mathcal{N}(x^{k-1}; \mu_\theta(x^k, k), \sigma_k^2 \mathbf{I}) + s\sigma_k^2 \nabla_{x^k} \log p(\mathbf{c}|x^k) \tag{32}$$

Here, $s\sigma_k^2 \nabla_{x^k} \log p(\mathbf{c}|x^k)$ is the guidance term that drives the generated image towards the target condition. $s$ is a scale parameter controlling the guidance strength. The gradient $\nabla_x \log p_k(\mathbf{c}|x)$ can be estimated in two ways: the first involves training an auxiliary model as described by [8] and the second method, described by Song *et al.* [51] and Luo [8], which doesn't require a auxiliary model. In Figure 9 the $\nabla_{x^k} \log p(\mathbf{c}|x^k)$ is termed $g(\mathbf{c}, x^k)$ as the guidance gradient.

Such guidance can also be applied to the continuous diffusion models as described by Song *et al.* [51]. For equation (19), it can be conditioned as:

$$\mathrm{d}x = \left\{ f(x, k) - g(k)^2 \left[ \nabla_x \log p_k(x) + \nabla_x \log p_k(\mathbf{c}|x) \right] \right\} \mathrm{d}k + g(k)d\bar{w} \tag{33}$$

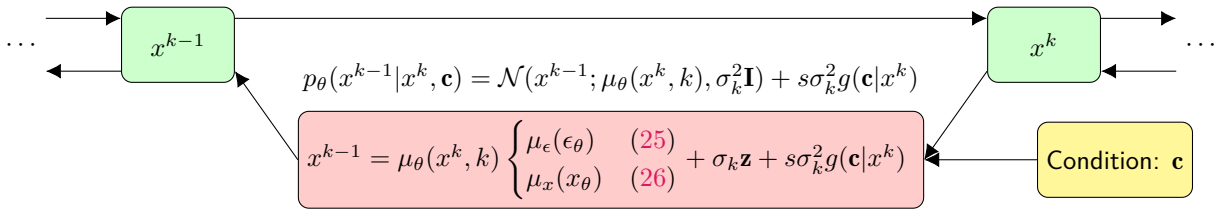

Figure 9: Illustration for conditional prediction models of the reverse process.

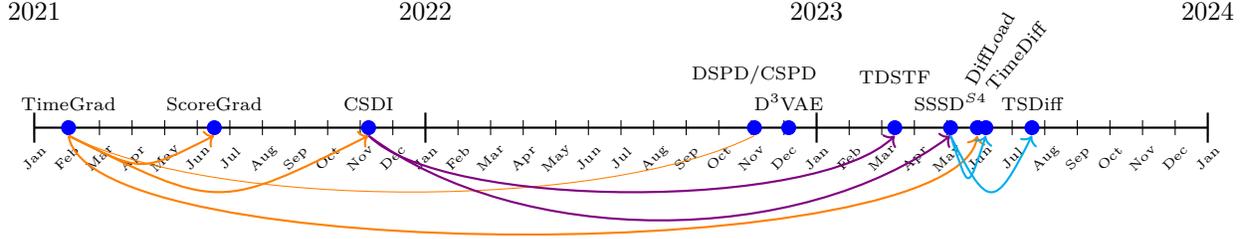

Figure 10: Diffusion for Time-series Forecasting Research Timeline

| Paper | Name | Method | Input Data | AR | Gen. Method | Rev. Model | Task | Horizon |
|---|---|---|---|---|---|---|---|---|
| [64] (3.1) | TimeGrad | Cond. Model | ← | ↻ | DDPM | Noise | 🔮 | Short |
| [62] (3.2) | ScoreGrad | Cond. Model | ← | ↻ | SDE | Score | 🔮 | Short |
| [65] (3.3) | CSDI | Cond. Model | ↔ | 🔨 | DDPM | Noise | 🩹 | Short |
| [66] (3.4) | DSPD | Cond. Model | ↔ | 🔨 | DDPM | Noise | 🩹 | Short |
| [66] (3.4) | CSPD | Cond. Model | ↔ | 🔨 | SDE | Noise | 🩹 | Short |
| [67] (3.5) | D$^3$VAE | Cond. Model | ← | 🔨 | Coupled DPM | BVAE | 🔮 | Short |
| [68] (3.6) | TDSTF | Cond. Model | ← | 🔨 | DDPM | Noise | 🔮 | Short |
| [69] (3.7) | SSSD$^{S4}$ | Cond. Model | ↔ | 🔨 | DDPM | Noise | 🩹 | Long |
| [17] (3.8) | DiffLoad | Cond. Model | ← | ↻ | DDPM | Noise | 🔮 | Short |
| [70] (3.9) | TimeDiff | Cond. Model | ← + Extras | 🔨 | DDPM | Data | 🔮 | Long |
| [71] (3.10) | TSDiff | Guidance | ← | 🔨 | DDPM | Noise | 🔮 | Short |

Table 1: Comparison of diffusion models for time-series prediction. **Method**: Conditioned Model (2.3.1), Guidance (2.3.2); **Input Data**: ← Data Before Prediction, ↔ Data Before and After Prediction; **AR**: ↻ Autoregressive, 🔨 Generated at Once (Non-AR); **Generative Method**: DDPM (2.1), SDE (2.2) **Reverse Model**: Noise (16), Data (17), Score (27); **Task**: 🔮 Forecasting, 🩹 Imputation. **Horizon**: Whether evaluated for Short ($\leq 50$ steps) or Long ($\geq 50$ steps) horizons.

## 3 Diffusion Models for Time Series Modeling

Time-series modeling using diffusion models has resurrected from sound generation via the methods WaveNet [63] and DiffWave [40], with the pioneering implementation named TimeGrad by Rasul *et al.* [64]. From the TimeGrad paper, many more followed as illustrated in Figure 10, and will be layed out in this chapter. First a brief introduction to the specifics of conditioning diffusion models for time-series forecasting is given. The conditioning of diffusion models for time-series can be done by conditioning the generation of the forecast through the reverse process in Equation 24 on the historical data like so:

$$p_\theta(X_{tar}^{0:K}|X_{obs}) = p(X_{tar}^K) \prod_{k=1}^{K} p_\theta(X_{tar}^{k-1} \mid X_{tar}^k, X_{obs}) \text{ where } p(X_{tar}^K) \sim \mathcal{N}(0, \mathbf{I}) \qquad (34)$$

The probability of $p_\theta(X_{tar}^{k-1} \mid X_{tar}^k, X_{obs})$ can be implemented through a conditional model as described in 2.3.1 or guidance in 2.3.2 with corresponding loss functions. The conditioning is not the historical data directly, but an encoding with the use of RNNs or Transformers [21] as will be evident in the following sections. In Table 1 an overview is presented of each of the implementations and their characteristics. In each section discussing a paper, the inspirational papers are mentioned that served as the foundation for the work. This is followed by an in depth but brief overview of the theory and intuition behind the models. Then the datasets and results are presented, finalizing with honest remarks and possible shortcomings are discussed.

| Dataset | DIM. | DOM. | FREQ. | # STEPS |
|---|---|---|---|---|
| Exchange | 8 | $\mathbb{R}^+$ | DAY | 6,071 |
| Solar | 137 | $\mathbb{R}^+$ | HOUR | 7,009 |
| Electricity | 370 | $\mathbb{R}^+$ | HOUR | 5,833 |
| Traffic | 963 | (0,1) | HOUR | 4,001 |
| Taxi | 1,214 | $\mathbb{N}$ | 30-MIN | 1,488 |
| Wikipedia | 2,000 | $\mathbb{N}$ | DAY | 792 |

Table 2: Gluon-TS Dataset Descriptions.

## 3.1 TimeGrad: Autoregressive Denoising Diffusion Models for Multivariate Probabilistic Time Series Forecasting (2021) [64]

TimeGrad represents the pioneer in using diffusion models for time-series forecasting, where its methodology lays in the DDPM [50]. The model architectures are based on those of WaveNet [63] and DiffWave [40].

TimeGrad was skillfully engineered to make probabilistic forecasts of the multivariate time-series, leveraging both past data and covariates. In this context covariates are extra conditions besides the historical data. Specifically, it employs an RNN mechanism to encode the time-series sequence, which, when combined with the DDPM, enables TimeGrad to proficiently learn the distribution of ensuing time steps. Given the use of the RNN, the model works by diffusing and denoising each multivariate value vector $x_t$ in an autoregressive manner. For this it encodes the historical data and covariates using the RNN starting from $x_{-H}$ and $\mathbf{h}_{-H} = 0$ as such

$$\mathbf{h}_t = \text{RNN}_\theta(\text{concat}(x_t^0, c_t), \mathbf{h}_{t-1}) \tag{35}$$

Where the reverse process as described in Equation 34 is done over the entire forecast, in TimeGrad it is done per time step.

$$p_\theta(X_{tar}^{0:K} | X_{obs}) = \prod_{t=1}^{F} p_\theta(x_t^{0:K} | \mathbf{h}_{t-1}) \tag{36}$$

Where $\mathbf{h}_0$ is the encoded historical data plus covaraites, furthermore, you can see, the conditioning $\mathbf{h}_{t-1}$ also evolves as the predictions go into the future. To continue, the paper models $\prod_{t=1}^{F} p_\theta(x_t^{0:K} | \mathbf{h}_{t-1})$ through the diffusion process, resulting in the extension of (36), which is in par with equation (34) as such:

$$p_\theta(X_{tar}^{0:K} | X_{obs}) = \prod_{t=1}^{F} p(x_t^K) \prod_{k=1}^{K} p_\theta(x_t^{k-1} | x_t^k, \mathbf{h}_{t-1}) \tag{37}$$

Because of this, the training objective is also extended to minimize the KL divergence over the entire forecast, which can be further simplified to with a noise prediction model:

$$\mathcal{L}_{t, \epsilon_\theta} = \mathbb{E}_{k, x_t^0, \epsilon} \left[ || \epsilon - \epsilon_\theta(x^k, k | \mathbf{h}_{t-1}) ||^2 \right] \tag{38}$$

$$\mathcal{L}_{\epsilon_\theta} = \frac{1}{F} \sum_{t=1}^{F} \mathcal{L}_{t, \epsilon_\theta} \tag{39}$$

The paper continuous to evaluate the model on six datasets. These datasets are preprocessed and stored by Salinas *et al.* [72][1] and can be easily accessed through GluonTS[2] and are described in Table 2. The number of historical steps used for conditioning is equal to the prediction length. The datasets and their original sources are Exchange [74], Solar [74], Electricity [75], Traffic [76], Taxi [77], and Wiki [78]. The results are displayed in Table 3.

While TimeGrad was recognized as state-of-the-art in multivariate probabilistic time-series forecasting at the time [64], it was not without limitations. Notably, its utilization of an RNN to encode historical time steps renders it challenging to apply to long-term time-series because it introduces the error accumulation issue [70]. As a recommendation Rasul *et al.* [64] suggest the use of Transformers [21]. Furthermore, The use of an RNN to encode the historical data and make future predictions makes it much slower compared to newer methods [70]. Overall, TimeGrad has served as the foundational paper for time-series forecasting, opening the doors for more diffusion research into this problem space.

---
[1] GP-Copula [72]: https://github.com/mbohlkeschneider/gluon-ts/tree/mv_release/datasets
[2] GluonTS: Probabilistic Time Series Models in Python [73] https://github.com/awslabs/gluonts

| Dataset | H | F | TimeGrad | | |
|---|---|---|---|---|---|
| | | | CRPS$_{\text{sum}}$ | CRPS | MSE |
| Exchange | 24 | 24 | 0.006±0.001 | 0.009±0.001 | 2.5e−4 |
| Solar | 24 | 24 | 0.287±0.020 | 0.367±0.001 | 8.8e2 |
| Electricity | 24 | 24 | 0.0206±0.001 | 0.049±0.002 | 1.97e5 |
| Traffic | 24 | 24 | 0.044±0.006 | 0.110±0.002 | 4.2e−4 |
| Taxi | 24 | 24 | 0.114±0.02 | 0.311±0.03 | 2.6e1 |
| Wikipedia | 30 | 30 | 0.0485±0.002 | 0.261±0.02 | 3.8e7 |

Table 3: TimeGrad Multivariate Forecasting results [64]

## 3.2 ScoreGrad: Multivariate Probabilistic Time Series Forecasting with Continuous Energy-based Generative Models (2021) [62]

Where TimeGrad [64] uses DDPM [50] for the diffusion process, ScoreGrad uses SDE [51] for the diffusion process as described in section 2.2. This shift to using SDE brings changes in terms of implementating the diffusion denoising processes, but the rest of the paper resembles that of TimeGrad. Just as TimeGrad the model is designed for forecasting tasks of multivariate time-series by diffusion and denoising single time-series vectors $x_t$, furthermore ScoreGrad also uses an RNN encode the historical data. Yan *et al.* [62] define this encoding with $\mathbf{F}_t$ and the update function as:

$$\mathbf{F}_t = R(\mathbf{F}_{t-1}, x_{t-1}(0), c_{t-1}) \tag{40}$$

This is much like the hidden state in TimeGrad, but the difference is that Yan *et al.* [62] leave the implementation up the user, suggesting the use of Temporal Convolutional Networks [63] or Transformers [21]. Just as the conditioning in (27), ScoreGrad conditions the reverse process with the encoding of $\mathbf{F}_t$ as follows:

$$dx_t = \left[ f(x_t, k) = g(k)^2 \nabla_{x_t} \log p_k(x_t|\mathbf{F}_t) \right] dk + g(k) d\bar{w} \tag{41}$$

With this extra conditioning and the autoregressive method, the loss function as shown in 28 changes to:

$$\mathcal{L}_{t,s_\theta} = \mathbb{E}_{k, x_t(0), x_t(k)} \left[ ||\nabla_{x_t(k)} \log p_{0k}(x_t(k)|x_t(0)) - s_\theta(x_t(k), k|\mathbf{F_t})||^2 \right] \tag{42}$$

Again, keep in mind that only a single value vector is diffused and denoised, and this is done for every future time step. As a result, the final loss function for the whole forecast is:

$$\mathcal{L}_{s_\theta} = \sum_{t=1}^{F} \mathcal{L}_{t,s_\theta} \tag{43}$$

The authors evaluate ScoreGrad on the same datasets accessed through GluonTS[2] as TimeGrad and compare with the CRPS$_{\text{sum}}$ evaluation metric. The results show that using the sub-VP SDE from equation (22) give the best results on all datasets apart from Electricity.

The results are a clear indication that the use of SDE for the diffusion process is superior compared to DDPM in this use-case. Furthermore, because ScoreGrad resembles TimeGrad so much, the remarks are similar too, such as for future works and as Yan *et al.* [62] already mentioned, it could benefit from the use of more advanced encoding models like Transformers [21] to extract time-series features. Furthermore, the forecasting is again autoregressive which causes errors to accumulate over the time horizon and is slow [70]. It is noteworthy that Yan *et al.* [62]'s paper omits specific details on the ODE application, as described in Equation 23. Nevertheless, their GitHub page briefly addresses this, highlighting that incorporating ODEs can significantly improve the sampling process, potentially increasing prediction speed by a factor of up to 4.9[3].

| Dataset | H | F | CRPS$_{\text{sum}}$ | | |
|---|---|---|---|---|---|
| | | | TimeGrad | VP SDE | sub-VP SDE |
| Exchange | 30 | 24 | 0.006±0.001 | 0.006±0.001 | 0.006±0.001 |
| Solar | 24 | 24 | 0.287±0.020 | 0.268±0.021 | **0.256±0.015** |
| Electricity | 24 | 24 | 0.0206±0.001 | **0.0192±0.001** | 0.0194±0.001 |
| Traffic | 24 | 24 | 0.044±0.006 | 0.043±0.004 | **0.041±0.004** |
| Taxi | 24 | 30 | 0.114±0.02 | 0.102±0.006 | **0.101±0.004** |
| Wikipedia | 30 | 24 | 0.0485±0.002 | **0.041±0.003** | 0.043±0.002 |

Table 4: ScoreGrad Multivariate Forecasting results [62]

## 3.3 CSDI: Conditional Score-based Diffusion Models for Probabilistic Time Series Imputation (2021) [65]

The CSDI implementation is much like that of TimeGrad [64], which is also based on DiffWave [40]. However instead of using a RNN to encode the conditional input, CSDI follows the future works of TimeGrad and makes use of a Transformer [21]. This modification allows the conditional input to be at arbitrary locations with respect to the target values, which allows imputation.

CSDI changes the problem statement to be compatible with imputation and does this by introducing a mask variable $m$ and a timestamp $s$ to the value vector, resulting in each time step resembling $\{x, m, s\}$. By introducing the mask, it can define which values are missing and require predictions. Keep in mind that forecasting is a specific case of imputation where all masked values are at the end of the time series sample. Furthermore, by introducing the timestamp, CSDI assumes that time intervals between consecutive data entries can be different allowing more flexibility in the data.

CSDI uses a slight modification of formalization of reverse process described in Equation 34, namely because of the imputation task, the *tar* and *obs* can be arbitrary points on the time-series and not necessarily split in historical and future points. Thus it becomes

$$p_\theta(X_{ta}^{0:K}|X_{co}) = p(X_{ta}^K) \prod_{k=1}^{K} p_\theta(X_{ta}^{k-1} \mid X_{ta}^k, X_{co}) \text{ where } p(X_{ta}^K) \sim \mathcal{N}(0, \mathbf{I}) \quad (44)$$

where $X_{ta}^{0:K}$ are the target timestamps and $X_{co}$ the conditional timestamps, future and past.

CSDI uses the same parameterization of DDPM and conditions the model itself as in Equation 25. This also leads to corresponding conditioned noise loss function

$$\mathcal{L}_{\epsilon_\theta} = \mathbb{E}_{k, X_{ta}^0, \epsilon} \left[ ||\epsilon - \epsilon_\theta(X_{ta}^k, k|X_{co})||^2 \right] \quad (45)$$

In CSDI, the same datasets are used as in TimeGrad and ScoreGrad but Exchange [74] is left out. The total number of steps as described in CSDI for the Traffic & Solar datasets are different. They should be 4001 and 7009 respectively to be the same as in TimeGrad & ScoreGrad through Gluon-TS[2]. These mistakes were verified through contact with the author on GitHub. Furthermore the historical window for the forecasting is increased, compared to Tables 3 and 4.

---
[3] https://github.com/yantijin/ScoreGradPred

| Dataset | H | F | CSDI ||| 
|---|---|---|---|---|---|
| | | | CRPS | CRPS$_{\text{sum}}$ | MSE |
| Solar | 168 | 24 | $0.338 \pm 0.012$ | $0.298 \pm 0.004$ | $9.0e2 \pm 6.1e1$ |
| Electricity | 168 | 24 | $0.041 \pm 0.000$ | $0.017 \pm 0.000$ | $1.1e5 \pm 2.8e3$ |
| Traffic | 168 | 24 | $0.073 \pm 0.000$ | $0.020 \pm 0.001$ | $3.5e-4 \pm 7.0e-7$ |
| Taxi | 48 | 24 | $0.271 \pm 0.001$ | $0.123 \pm 0.003$ | $1.7e1 \pm 6.8e-2$ |
| Wiki | 90 | 30 | $0.207 \pm 0.002$ | $0.047 \pm 0.003$ | $3.5e7 \pm 4.4e4$ |

Table 5: CSDI Multivariate Forecasting results [65]

This increase in the historical window length may serve as an advantage, the results of TimeGrad in the paper were taken from Rasul *et al.* [64] and thus did not have the same historical window length. This extended context could positively impact the results in CSDI's favour on top of the increase because of the architecture.

Furthermore, the method of masking used in CSDI is similar to inpainting in computer vision tasks. This inpainting has shown to cause artifacts between masked and observed regions call boundary disharmony [79]. The masking method used in CSDI is also empirically shown the have the same effects in time-series [70].

CSDI employs a automatic masking algorithm in order to generate the training data. This feature has a negative effect in situations where the data is extremely sparse as tackled in TDSTF by Chang *et al.* [68]. In this paper, CSDI is used as a baseline for sparse data and does not deal with such sparse situations by excluding enough invalid information and thus over 99.5% of the conditional data points were missing on average. This resulted in TDSTF being a order of magnitude faster than CSDI in training.

Furthermore the computation time of the denoising network is quadratic in the number of variables and number of time-series points because it is based on two transformer networks. This causes it to run out of memory easily when modeling long multivariate time-series [70].

Just like SSSD$^{S4}$ the conditioning is done in the denoising networks' intermediate layers and introduce inductive bias into the denoising objective, Shen and Kwok [70] hypothesises that this may not be enough to guide the network in capturing information from the history and leads to inaccurate predictions.

CSDI, along with TimeGrad, has played a crucial role in the development of imputation and forecasting diffusion models. Its widespread use as a baseline highlights its importance and establishes it as a pillar in the rise of diffusion models for time-series forecasting.

## 3.4 DSPD & CSPD: Modeling Temporal Data as Continuous Functions with Process Diffusion (2022) [66]

The implementation is based on that of TimeGrad [64] with two key contributions. The main contribution and novelty of this paper is that they model the time-series as continuous functions instead of discrete sequences. This allows the model to predict values at any time, even between the existing time-series data steps. The second contribution is a change in the architecture that allows the model to predict multiple values at the same time which scales better on modern hardware.

Formally, they make the assumption that each observed time series comes from an underlying continuous function $x(\cdot)$ and thus are interested in modeling the distribution $p(x(\cdot))$. Because of this assumption, they have to add a time correlated noise function $\epsilon(\cdot)$ to $x(\cdot)$ instead of independent noise. As noise functions they propose a Gaussian process prior (GP) or a Ornstein-Uhlenback (OU) diffusion process. Biloš *et al.* [66] define both the GP and OU processes with a multivariate normal distribution $\mathcal{N}(0, \Sigma)$. Both $\epsilon(\cdot)$ and $\Sigma$ are calculated using the timestamps of the observations.

They continue by proposing their implementations Discrete Stochastic Process Diffusion (DSPD) and Continuous Stochastic Process Diffusion (CSPD), based on DDPM and SDE respectively. The fact that the implementations model continuous functions has nothing to do with whether the diffusion process is done discretely or continuously. In both the DSPD and CSPD, they use a forecasting model that outputs noise $\epsilon$ as in Equation 25. For CSPD, they refactor the noise predictions to the score function value.

For the case of forecasting, an RNN is used to obtain the historical condition **z**. This condition does not change during the forecasting as poposed to TimeGrad and ScoreGrad. Furthermore, because the architecture uses 2D instead of 1D convolutional layers, the model is capable of forecasting all of the values at once.

**Discrete Stochastic Process Diffusion (DSPD):** In the reverse process the marginal distributions, as in Equation 24 are adjust for the assumption of modeling continuous functions as such:

$$p_\theta(X_{tar}^{k-1} \mid X_{tar}^k, X_{obs}) = \mathcal{N}(x^{k-1}; \mu_\theta(x^k, t, k|\mathbf{z}), \sigma_k^2 \Sigma) \tag{46}$$

where $t$ are the timestamps of the to be forecasted values and **z** the encoded historical data. Furthermore, the variance is adjusted to the time correlated noise with $\Sigma$ instead of **I** as in Equation 9.

**Continuous Stochastic Process Diffusion (CSPD):** Given the factorized covariance matrix $\Sigma = \mathbf{LL}^T$ the variance preserving SDE of Equation 21 is modified to:

$$dX_{tar}^k = -\frac{1}{2}\beta(k) X dk + \sqrt{\beta(k)} \mathbf{L} dW \tag{47}$$

For both DPSD and CSPD, the authors employ the reparameterization to predict noise which also results in the following objective function:

$$\mathcal{L}_{\epsilon_\theta} = \mathbb{E}_{k, X_{tar}^0, \epsilon} \left[ ||\epsilon - \epsilon_\theta(X_{tar}^k, t, k, |\mathbf{z})||^2 \right] \tag{48}$$

The paper does not compare with the same CRPS$_{\text{sum}}$ metric as previous papers, but it does state that the order compared to TimeGrad does not change. Instead the comparison is done with metrics NRMSE and the energy score [80].

| Dataset | H | F | NRMSE | | Energy Score | |
|---|---|---|---|---|---|---|
| | | | TimeGrad | DSPD-GP | TimeGrad | DSPD-GP |
| Electricity | 24 | 24 | $0.064 \pm 0.007$ | $\mathbf{0.045 \pm 0.002}$ | $8425 \pm 613$ | $\mathbf{7079 \pm 164}$ |
| Exchange | 30 | 30 | $0.013 \pm 0.003$ | $\mathbf{0.012 \pm 0.001}$ | $0.057 \pm 0.002$ | $\mathbf{0.031 \pm 0.002}$ |
| Solar | 24 | 24 | $0.799 \pm 0.096$ | $\mathbf{0.757 \pm 0.026}$ | $\mathbf{150 \pm 17}$ | $166 \pm 12$ |

Table 6: DSPD-GP Multivariate Forecasting results [66]

The process has the same limitation as TimeGrad[3.1] and ScoreGrad [3.2], namely that it uses an RNN to represent the historical data when evaluated for forecasting. For imputation, the models use the same architecture with Transformer, as CSDI [3.3] and only changes the noise source to that of a Gaussian process. The paper only compares to TimeGrad for forecasting capabilities and compares the imputation capabilities with CSDI. A forecasting comparison with CSDI would have been interesting as it already performs better than TimeGrad at forecasting. The code is not available for reproducibility at the moment of writing this survey. The paper could make a stronger case for the superiority of time correlated noise if they had evaluated it against more time-series forecasting models and used Transformers for encoding the historical data as these possible improvements were already presented in the imputation part of the paper.

## 3.5 D$^3$ VAE: Generative Time Series Forecasting with Diffusion, Denoise, and Disentanglement (2022) [67]

This paper, even though it mentions the diffusion and denoising process of Ho *et al.* [50], it does so differently from what has been described so far in this survey. Furthermore, the paper focuses on the interpretability of the forecasts and tries to decompose the aleatoric and epistemic uncertainty of the data and model. The model contains three key components. Firstly a coupled diffusion process, secondly a bidirectional variational auto-encoder(BVAE) [81] for disentanglement, and lastly extra denoising through denoising score matching (DSM). These three components also motivate the name with diffusion denoise and disentanglement, hence D$^3$ VAE.

**Coupled Diffusion Process:** The name, coupled diffusion process, comes from the fact that it adds noise to both the historical and future windows of a sample. This helps because the diffused samples can augment the dataset giving more support to forecasting with short and noisy time series. Furthermore, the paper proves that by decomposing the $X_{tar}^k$ from Equation 8 and $X_{obs}^k$, into an ideal part and noisy part, it can reduce the difference between diffusion noise and generation noise. The decomposition is formalized as:

$$X_{tar}^k = \sqrt{\alpha_k'}X_{tar}^0 + \sqrt{1-\alpha_k'}\epsilon_{X_{tar}} = \underbrace{\sqrt{\alpha_k'}\bar{X}_{tar}}_{\text{ideal part}} + \underbrace{\sqrt{\alpha_k'}\delta_{X_{tar}} + \sqrt{1-\alpha_k'}\epsilon_{X_{tar}}}_{\text{noisy part}} \quad (49)$$

$$X_{obs}^k = \sqrt{\alpha_k}X_{obs}^0 + \sqrt{1-\alpha_k}\epsilon_{X_{obs}} = \underbrace{\sqrt{\alpha_k}\bar{X}_{obs}}_{\text{ideal part}} + \underbrace{\sqrt{\alpha_k}\delta_{X_{obs}} + \sqrt{1-\alpha_k}\epsilon_{X_{obs}}}_{\text{noisy part}} \quad (50)$$

where the diffusion of the targets uses a scaled version of $\alpha_k$, namely given a scale parameter $\omega \in (0,1)$ such that $\beta_k' = \omega\beta_k$ and subsequently $\alpha_k' = \prod_{i=1}^{k}(1-\beta_i')$. The noisy part is split into the diffusion noise $\epsilon_X$ and data noise $\delta_X$. In other words, this decomposition reduces the uncertainty of the generated forecasts as proven by Li *et al.* [67].

**Bidirectional Variational Auto-Encoder:** The second component is the more efficient generative model, i.e. BVAE [81], that replaces the reverse process. Besides replacing the reverse process the BVAE is also used to facilitate model interpretability through disentanglement. In contrast to the standard reverse process that diffuses and denoises only the targets, the BVAE processes the diffused observations $X_{obs}^k$. Formally, the BVAE encodes the $X_{obs}^k$ into a latent variable $Z$ as $p_\phi(Z, X_{obs}^k)$. Given this latent variable the decoder generates the prediction $\hat{X}_{tar}^k$ through $p_\theta(\hat{X}_{tar}^k, Z)$. The generated prediction still contains noise, which is subsequently removed by the Scaled DSM. The Latent variable $Z$ is used for disentanglement, which is achieved by minimizing the Total Correlation [82, 83]. This helps with interpretability by identifying the independent factors of the data [84, 85, 86].

**Scaled Denoising Score Matching:** The forecasts produced by the BVAE tend to move towards the diffused target series. To further clean the generated target series $\hat{X}_{tar}^k$, the authors employ the DSM to accelerate the de-uncertainty process without sacrificing the model flexibility. This denoising is done via a single-step gradient jump [87].

$$\hat{X}_{tar}^0 = \hat{X}_{tar}^k - \sigma_0^2 \nabla_{\hat{X}_{tar}^k} E(\hat{X}_{tar}^k, \zeta) \quad (51)$$

where $\sigma_0$ is a hyperparameter and $E(\hat{X}_{tar}^k, \zeta)$ is an energy function. The denoising step is taken in the direction of the greatest increase of this energy function, hence $\nabla_{\hat{X}_{tar}^k}$.

All of these components are trained using their own loss functions, which results in the total loss function:

$$\mathcal{L} = \underbrace{\psi \cdot \mathcal{D}_{KL}\left(q(X_{tar}^k||p_\theta(X_{tar}^k)\right)}_{BVAE} + \underbrace{\lambda \cdot \mathcal{L}(\zeta, k)}_{DSM} + \underbrace{\gamma \cdot \mathcal{L}_{TC}}_{Disent.} + \underbrace{\mathcal{L}_{MSE}(\hat{X}_{tar}^k, X_{tar}^k)}_{BVAE} \quad (52)$$

D³VAE use the datasets: Traffic [76][4], Electricity [75], Weather [88], ETTm1[24][5], ETTh1[24][5], Wind[67][6]. It is not explained how the datasets of Electricity [75] and Weather [88] are gathered or pre-processed, only the source websites are mentioned. It is important to note that the Traffic and Electricity datasets used in D³VAE do not share characteristics with those used from GluonTS[2] in TimeGrad, ScoreGrad or CSDI even though they are from the same source. Furthermore and because of the motivation to be able to train on short time-series datasets via augmentation, the authors only take a small percentage of the datasets previously mentioned. The results are presented in Table 7 with the percentage used behind the dataset name.

There is a big difference in MSE and CRPS between D³VAE and TimeGrad. Notably in some datasets, this difference reaches up to two orders of magnitude. In the D³VAE paper are plots of the TimeGrad predictions, which show an almost binary pattern, suggesting the model is unable to predict any useful information. Even though that the pre-processing is not explained by D³VAE, the source of the Electricty dataset is the same as in TimeGrad [64]. An more in depth analysis into these differences and their underlying causes would be beneficial. It could offer valuable insights into the model behavior and enhance the understanding of these divergent outcomes. The authors of TimeDiff 3.9 have used D³VAE in their comparison and the results show for multivariate time-series forecasting with long forecasting windows that D³VAE performs better than TimeGrad and CSDI [70]. The paper claims the BVAE replaces the reverse process while the inputs to the BVAE is the observation section of the time-series instead of the future section. This makes the diffusion process untypical and more difficult [70]. Furthermore, the authors claim that the outputs of the BVAE converge to the diffused outputs of the coupled diffusion process. The outputs of the coupled diffusion process are maximum noise, indicating that the BVAE tends to move towards that noisy output. Moreover, the final denoising step is taken by the DSM creating the cleaned-up prediction. The model has many hyperparameters that need to be tuned, this can be a significant limitation in finding its optimal performance.

---

[4] LSTNet [74]: https://github.com/laiguokun/multivariate-time-series-data/tree/master
[5] Informer [24]: https://github.com/zhouhaoyi/ETDataset
[6] D³VAE [67]: https://github.com/PaddlePaddle/PaddleSpatial/tree/main/paddlespatial/datasets

| Dataset | H | F | MSE | | CRPS | |
|---|---|---|---|---|---|---|
| | | | **D3VAE** | TimeGrad | **D3VAE** | TimeGrad |
| Traffic (5%) | 8 | 8 | 0.081 ± 0.003 | 3.695 ± 0.246 | 0.207 ± 0.003 | 1.410 ± 0.027 |
| | 16 | 16 | 0.081 ± 0.009 | 3.495 ± 0.362 | 0.200 ± 0.014 | 1.329 ± 0.057 |
| | 32 | 32 | 0.091 ± 0.007 | 5.195 ± 2.26 | 0.216 ± 0.012 | 1.565 ± 0.329 |
| | 64 | 64 | 0.125 ± 0.005 | 3.692 ± 1.54 | 0.244 ± 0.006 | 1.412 ± 0.257 |
| Electricity (3%) | 8 | 8 | 0.251 ± 0.015 | 2.703 ± 0.087 | 0.398 ± 0.011 | 1.208 ± 0.024 |
| | 16 | 16 | 0.308 ± 0.030 | 2.770 ± 0.237 | 0.437 ± 0.020 | 1.240 ± 0.048 |
| | 32 | 32 | 0.410 ± 0.075 | 2.640 ± 0.138 | 0.534 ± 0.058 | 1.234 ± 0.027 |
| Weather (2%) | 8 | 8 | 0.169 ± 0.022 | 1.110 ± 0.083 | 0.357 ± 0.024 | 0.733 ± 0.016 |
| | 16 | 16 | 0.187 ± 0.047 | 1.065 ± 0.145 | 0.361 ± 0.046 | 0.724 ± 0.021 |
| | 32 | 32 | 0.203 ± 0.008 | 1.178 ± 0.069 | 0.383 ± 0.007 | 0.696 ± 0.011 |
| | 64 | 64 | 0.191 ± 0.022 | 1.063 ± 0.061 | 0.358 ± 0.044 | 0.696 ± 0.011 |
| ETTm1 (1%) | 8 | 8 | 0.527 ± 0.073 | 0.984 ± 0.074 | 0.557 ± 0.048 | 0.908 ± 0.038 |
| | 16 | 16 | 0.968 ± 0.104 | 2.032 ± 0.234 | 0.821 ± 0.072 | 0.919 ± 0.031 |
| | 32 | 32 | 0.707 ± 0.061 | 1.251 ± 0.133 | 0.697 ± 0.040 | 0.822 ± 0.032 |
| ETTh1 (5%) | 8 | 8 | 0.292 ± 0.036 | 4.259 ± 1.13 | 0.424 ± 0.033 | 1.092 ± 0.028 |
| | 16 | 16 | 0.374 ± 0.061 | 1.332 ± 0.125 | 0.488 ± 0.039 | 0.879 ± 0.037 |
| | 32 | 32 | 0.334 ± 0.008 | 1.514 ± 0.042 | 0.461 ± 0.004 | 0.925 ± 0.016 |
| | 64 | 64 | 0.349 ± 0.039 | 1.150 ± 0.118 | 0.473 ± 0.024 | 0.835 ± 0.045 |
| Wind (2%) | 8 | 8 | 0.681 ± 0.075 | 12.67 ± 1.75 | 0.596 ± 0.052 | 1.440 ± 0.059 |
| | 16 | 16 | 1.033 ± 0.062 | 12.86 ± 2.60 | 0.757 ± 0.053 | 1.240 ± 0.070 |
| | 32 | 32 | 1.224 ± 0.060 | 13.10 ± 0.955 | 0.869 ± 0.074 | 1.518 ± 0.020 |
| | 64 | 64 | 0.902 ± 0.024 | 3.857 ± 0.597 | 0.761 ± 0.021 | 1.110 ± 0.143 |
| Electricity (100%) | 16 | 16 | 0.330 ± 0.033 | 46.69 ± 3.13 | 0.445 ± 0.020 | 2.702 ± 0.79 |
| | 32 | 32 | 0.336 ± 0.017 | 30.94 ± 1.70 | 0.444 ± 0.015 | 2.476 ± 0.042 |
| ETTm1 (100%) | 16 | 16 | 0.018 ± 0.002 | 68.26 ± 2.04 | 0.102 ± 0.003 | 1.153 ± 0.019 |
| | 32 | 32 | 0.034 ± 0.001 | 53.47 ± 26.1 | 0.144 ± 0.006 | 1.083 ± 0.109 |

Table 7: D$^3$VAE Multivariate forecasting results [67]

## 3.6 TDSTF: Transformer-based Diffusion probabilistic model for Sparse Time series Forecasting (2023) [68]

The TDSTF model, drawing inspiration from CSDI [65], is specifically designed for forecasting ICU patient vital signs using sparse historical data. Addressing the challenge of sparse multivariate ICU data, the model introduces key innovations. Its main contribution is efficiently representing the spare data with a triplet form, which contains a feature, time, and value. On top of that, there is also a mask bit indicating the presence or absence of data of a triplet. This compact representation effectively stores sparse data, maintaining the integrity of temporal information and reducing the noise that typically arises from imputation or aggregation methods. Furthermore, as the name implies, a Transformer [21] is used to encode the historical data. There are also adjustments in selecting the inputs to the encoder. First of all, if the mask indicates the absence of a triplet, it won't be included to the Transformer. Furthermore, if there are too many triplets as input to the model, only the data points of the same target feature, or most correlated to the target data are included. This ensures that the most relevant information is included in the model's input.

The reverse process is formalized as in Equation 34 with a noise prediction model and corresponding loss function

$$\mathcal{L}_{\epsilon_\theta} = \mathbb{E}_{k,X_{tar}^0,\epsilon} \left[ ||\epsilon - \epsilon_\theta(X_{tar}^k, k|X_{obs})||^2 \right]. \tag{53}$$

The ICU data contains the vital signs Heart Rate (HR), Systolic Blood Pressure (SBP), and Diastolic Blood Pressure (DBP). Furthermore, it also contains special events that occur with the patient such as administering drugs. This approach allows for a more accurate forecast by considering the interrelated events, interventions, and conditions that could impact the patient. The paper states that 60 triplets are used to encode all of the historical data, however it is not mentioned how many steps in the future the vital signs are predicted. The paper merely states it makes a 10-minute forecast of the three vital signs HR, SBP, and DBP.

The specific data set is MIMIC-III [89][7] and the performance are evaluated with MSE and NACRPS as described in section 1.1.2. The results are presented in Table 8.

| Dataset | H | F | NACRPS | | MSE | |
|---|---|---|---|---|---|---|
| | | | CSDI | TDSTF | CSDI | TDSTF |
| MIMIC-III | 60tr | 10m | $0.5470 \pm 0.0040$ | $\mathbf{0.4438 \pm 0.0078}$ | $0.6341 \pm 0.0098$ | $\mathbf{0.4168 \pm 0.0232}$ |

Table 8: TDSTF ICU Vital Signs Forecasting results [68]

Besides achieving better predictive results, TSDTF is also much faster to train and inference compared to CSDI [3.3]. In the experiments, CSDI has 280 thousand parameters and TDSTF 560 thousand, even with this increase, TDSTF only took 30 minutes to train compared to the 12 hours for CSDI. Furthermore, CSDI took 14.913 seconds to inference while TDSTF only took 0.865 seconds, making TDSTF more than 17 times faster. TSDTF only shown to be an improvement over CSDI for the sparse dataset MIMIC-III[7]. More insights in its performance could be achieved by testing it against the same datasets mentioned in the CSDI paper [65]. Overall the paper makes a strong case, solving the shortcomings of CSDI when it comes to sparse data.

## 3.7 SSSD$^{S4}$: Diffusion-based Time Series Imputation and Forecasting with Structured State Space Models (2023) [69]

The model of SSSD$^{S4}$, is heavily inspired by DiffWave [40] and CSDI [65], with three main differences. The first is the fact that SSSD$^{S4}$ uses S4 models [90] instead of dilated convolutions or transformer layers [21]. The S4 models are computationally more efficient and particularly suited to handling long-term dependencies in time-series data [90, 91]. Secondly, SSSD$^{S4}$ has reduced the internal representation of shape (batch *dim*, diffusion *dim*, input channel *dim*, time *dim*) in CSDI to shape (batch *dim*, diffusion *dim*, time *dim*). This maps the input channels into the diffusion dimension and thus the model only performs diffusion along the time dimension instead of both the time and input channel dimensions. This corresponds to the common approach when applying diffusion models to image or sound data [69].

At last, SSSD$^{S4}$ only applies the diffusion process to the segments that need to be imputed, as this yields better results than applying the diffusion process to all segments. This change in applying the noise during the diffusion process turns out to be better than the methods in image inpainting [79]. This is different from CSDI, which applies noise to the imputation targets and missing values. The missing values in CSDI are padded with zeros to fix the shape of the inputs. Whereas in SSSD$^{S4}$ no noise is generated for the missing values.

---
[7] TDSTF [68]: https://github.com/PingChang818/TDSTF

The input data is represented differently from CSDI, leaving out the timestamp. This results in $\{x, m\}$ pairs where $x$ is the value vector and $m$ is the binary mask value.

The reverse process and loss function are formulated the same as in Equation 44 and Equation 45. The authors investigate both types of conditioning, as extra input during training and training unconditionally and including the conditional information during inference.

**Structured State Space Sequence Model (S4)**: The S4 architecture is composed of State Space Models (SSM). The SSM [90] is based on a linear state space transition equation, connecting a 1D input sequence $u(t)$ to a 1D output sequence $y(t)$ via a N-dimensional hidden state $x(t)$. This can be formalized as

$$x'(t) = Ax(t) + Bu(t) \text{ and } y(t) = Cx(t) + Du(t), \tag{54}$$

where $A$, $B$, $C$, $D$ area transition matrices. These SSMs can be evaluated efficiently on modern GPUs [90] and can capture long-term dependencies according to the HiPPO theory [92, 93]. By stacking several SSM blocks and additional layers the Structured State Space Sequence model (S4) was created and demonstrated good performance on sequence classification tasks. Structuring the S4 layers in a U-Net-inspired configuration gave rise to SaShiMi [91], which is a generative sequence generative model.

Because the model is inspired by DiffWave and CSDI, the authors propose to compare 3 variations of models to better understand each of the individual components. Namely, $SSSD^{S4}$, $SSSD^{SA}$, $CSDI^{S4}$. $SSSD^{S4}$ is a variant of DiffWave but with S4 components. $SSSD^{SA}$ is a variant of DiffWave but with the SaShiMi instead of S4 components. Lastly, $CSDI^{S4}$ is a variant of CSDI but where the time direction transformer is replaced with an S4 layer.

$SSSD^{S4}$ performs time-series forecasting on the PTB-XL [94], the Solar [74] dataset from Gluon-TS[2] and on the ETTm1[5] dataset from Zhou *et al.* [24]. All data gathering and pre-processing for $SSSD^{S4}$ is explained on their Github[8].

| Dataset | H | F | MAE | | | | MSE | |
|---|---|---|---|---|---|---|---|---|
| | | | CSDI | $CSDI^{S4}$ | $SSSD^{SA}$ | $SSSD^{S4}$ | CSDI | $SSSD^{S4}$ |
| PTB-XL | 800 | 200 | 0.165±9e−4 | 0.120±2e−4 | **0.087**±8e−3 | 0.090±3e−3 | - | - |
| Solar | 168 | 24 | - | - | - | - | 900±61 | **503**±10.6 |
| ETTm1 | 96 | 24 | 0.370 | - | - | **0.361** | 0.354 | **0.351** |
| | 48 | 48 | 0.546 | - | - | **0.479** | 0.750 | **0.612** |
| | 284 | 96 | 0.756 | - | - | **0.547** | 1.468 | **0.538** |
| | 288 | 288 | **0.530** | - | - | 0.648 | **0.608** | 0.797 |
| | 384 | 672 | 0.891 | - | - | **0.783** | 0.946 | **0.804** |

Table 9: $SSSD^{S4}$ Multivariate Forecasting Results [69]

On the PTB-XL dataset, it can be seen that $SSSD^{SA}$ and $SSSD^{S4}$ perform similarly well for forecasting. However, for imputation on PTB-XL the S4 variant outperforms SA all the time. Furthermore, in the visual plots shown by [69] the S4 variant is much more certain in its predictions. The comparison of CSDI [3.3] and $SSSD^{S4}$ on Solar indicates a great improvement, just as on ETTm1. There are however also limitations to the $SSSD^{S4}$ method. As hypothesized in the paper, because of the dimension reduction, it becomes increasingly difficult for the $SSSD^{S4}$ model to reconstruct the original input channels from the internal diffusion channels in situations with a large number of input channels. Alcaraz and Strodthoff [69] see the central strength of $SSSD^{S4}$ to be its ability to capture long-term dependencies along the time direction. As a result, CSDI performs better in situations with many input channels or situations where the relations between input channels are more important than the temporal consistency [69]. The conditioning strategy is, like CSDI, taken from models for image or text data, and not specifically for time series. It has been empirically shown that its long-range prediction performance is inferior to other time-series prediction models [70]. Furthermore, The performance deteriorated due to the boundary disharmony by the semi-supervised learning strategy [70]. The research in SSMs is continuing at a high pace, surpassing Transformers in foundation models such as [95].

---

[8] $SSSD^{S4}$ [69]: https://github.com/AI4HealthUOL/SSSD

## 3.8 DiffLoad: Uncertainty Quantification in Load Forecasting with Diffusion Model (2023) [17]

The model mentions TimeGrad [64] but in essence is a sequence-to-sequence model [20] that performs the diffusion denoising process on the hidden state in between the encoder and decoder. It is specifically designed for forecasting tasks of electrical loads and to tackle situations where the temporal data has a distribution shift and contains outliers. The authors make two main contributions, the first one being a new uncertainty quantification method for neural network forecasting utilizing a diffusion-based encoder to concentrate uncertainty in the latent variable before inputting it into the decoder, providing better insights into the epistemic uncertainty. Furthermore, they propose an emission head based on the additive Cauchy distribution to capture the aleatoric uncertainty.

**Hidden-state Diffusion:** The historical data is encoded with the use of a GRU, resulting in a hidden state $h_0$. This hidden state is then put through the diffusion and denoising process of an unconditional DDPM [2.1] with $p(h^K) \sim \mathcal{N}(0, \mathbf{I})$ as:

$$p_\theta(h^{0:K}) = p(h^K) \prod_{k=1}^{K} p_\theta(h^{k-1} \mid h^k) \text{ where } p_\theta(h^{k-1}|h^k) = \mathcal{N}(h^{k-1}; \mu_\theta(h^k, k), \sigma_k^2 \mathbf{I}) \tag{55}$$

The theory continuous with Equation 10 where $\sigma_k^2 = \tilde{\beta}_k$ and $\mu_\theta(h^k, k)$ is reparameterized with the noise prediction model explained in Equation 16 with loss function $\mathcal{L}_{\epsilon_\theta}$. The idea behind this diffusion process is to be able to capture and quantify the epistemic uncertainty by generating multiple forecasts, from which an uncertainty can be determined. By estimating the distribution of the hidden state of encoded historical data, the randomness that is left in the estimated hidden state quantifies the model uncertainty.

**Cauchy distribution model head:** The new hidden state $\hat{h}_0$ is used as the starting hidden state for the decoder and it continuous to produce the forecast by outputting $\hat{\mu}_\phi$ and $\hat{\sigma}_\phi$ for the Cauchy distribution. The Cauchy distribution is used because it is more heavy-tailed compared to the traditional Gaussian distribution. This heavy-tail property makes the Cauchy distribution more robust to outliers and mutation [96, 97]. The Cauchy distribution can be defined by a mean and scale parameter, similar to the Gaussian distribution and is formalized by the authors as:

$$\mu_{\phi(t+1)} = \text{NN}_1(\hat{h}_{t+1}), \tag{56}$$

$$\sigma_{\phi(t+1)} = \text{SoftPlus}\left[\text{NN}_2(\hat{h}_{t+1})\right] \tag{57}$$

Using these parameters, the conditional distribution of the error can be described with the Cauchy distribution $\mathcal{C}$:

$$p_\phi(x_{t+1}|\hat{h}_{h+1} = \mathcal{C}(x_{t+1}; \mu_{\phi(t+1)}, \sigma_{\phi(t+1)}) \tag{58}$$

The loss for the decoder and the diffusion model are combined as such

$$\mathcal{L} = \lambda \cdot \mathcal{L}_{\epsilon_\theta} - \log \sigma_\phi + \log (X_{tar} - \mu_\phi)^2 + \sigma_\phi^2 \tag{59}$$

During inference the model makes $M$ predictions. These predictions have an uncertainty, because of the probabilistic nature of the diffusion model, which can be described as a Gaussian distribution with $\sigma_\theta$. This Gaussian distribution represents the epistemic uncertainty. The uncertainties outputted through the Cauchy head can be regarded as the aleatoric uncertainty. Because both uncertainties are $\alpha$-stable they can be added together to form the final uncertainty of the prediction.

$$\sigma = \underbrace{\sigma_\phi}_{aleatoric} + \underbrace{\sigma_\theta}_{epistemic} \tag{60}$$

Authors continue to evaluate the model on 3 datasets, Global Energy Forecasting (GEF) [98], Building Data Genome Project 2 (BDG2) [99], and Day-ahead electricity demand forecasting (COV) [100]. The authors performed an ablation study to measure the effects of the model as can be seen in Table 10. The o/o variant is without diffusion and Gaussian distribution heads on the decoder. The d/o variant is with diffusion and Gaussian, and d/c is with diffusion and Cauchy distribution. The results show a clear superiority to the model with diffusion and Cauchy distribution.

| Dataset | H | F | MAE | | | CRPS | | |
|---------|---|---|-----|-----|-----|------|-----|-----|
|         |   |   | o/o | d/o | d/c | o/o | d/o | d/c |
| GEF  | - | - | 119.47 | 115.37 | **110.44** | 86.61 | **83.07** | 83.24 |
| COV  | - | - | 21844.0 | 22512.37 | **20308.3** | 16263.42 | 16705.65 | **15435.79** |
| BDG2 | - | - | 12.76 | 12.38 | **11.18** | 9.35 | 9.16 | **8.72** |

Table 10: DiffLoad Forecasting results [17]

The sequence-to-sequence model implemented with RNN served as a pillar in academia, is now being overshadowed by more advanced variants that incorporate Transformer layers [28]. There is a lack of theoretical background in the authors' discussion on differentiating between epistemic and aleatoric uncertainties. This is mainly evidenced through enhanced prediction performance, yet the process by which the model identifies these uncertainties remains unexplained. Furthermore, it is not mentioned what the historical and future lengths are making it difficult to reproduce the results. This is the only implementation that performs the diffusion on a latent variable and not on the time-series directly [35], however, it is not typically conditioned latent space diffusion because the encoder's output is the "historical condition" for the decoder to generate the forecasts.

### 3.9 TimeDiff: Non-autoregressive Conditional Diffusion Models for Time Series Prediction (2023) [70]

The TimeDiff model addresses the shortcomings of CSDI and SSSD$^{S4}$ by introducing additional inductive bias in the conditioning module that is tailor-made for time-series. It does this by two conditioning mechanisms: "Future Mixup" and "Autoregressive initialization". Future mixup is based on mixup by Zhang *et al.* [101] and is extended to randomly reveal parts of the ground-truth future predictions during training. Autoregressive initialization works by learning a linear autoregressive model on the data that provides an initial linear guess for the predictions.

**Future mixup:** The ideal condition generates the forecast is $X_{tar}$ itself, this is not available during inference, but it is during training. Inspired by mixup [101], the revealing of the ground-truth works by creating a conditioning variable $z_{mix} \in \mathbb{R}^{d \times F}$ that is of the same size as $X_{tar}$. During inference this variable will be filled with a mapping of the historical data as seen in Equation 61. However, since this historical data can have a different size it is mapped through a convolution network $\mathcal{F}$ to end up with the same dimensions.

$$z_{mix} = \mathcal{F}(X_{obs}) \tag{61}$$

Now, during training, randomly selected data points of $X_{tar}$ are also included, as seen in Equation 62, by using a mask variable $m^k \in [0,1)^{d \times F}$ where each value is sampled from the uniform distribution $[0,1)$.

$$z_{mix} = m^k \odot \mathcal{F}(X_{obs}) + (1 - m^k) \odot X_{tar}^0 \tag{62}$$

**Autoregressive initialization:** For non-autoregressive models often produce disharmony at the boundaries between masked and observed regions [79]. Shen and Kwok [70] state that for time series forecasting, this translates to disharmony between historical and future segments. To solve this problem they propose the linear autoregressive (AR) model $\mathcal{M}_{ar}$ which will make an initial guess $z_{ar} \in \mathbb{R}^{d \times H}$ for $X_{tar}$. This model cannot capture the complex non-linear patterns in time-series, but it can approximate the simple ones, such as short-term trends [74]. The initial guess $z_{ar}$ is made by training the following to predict $X_{tar}$:

$$z_{ar} = \sum_{t=-H}^{0} W_t \odot \bar{X}_t + B \tag{63}$$

where $\bar{X}_t^0 \in \mathbb{R}^{d \times F}$ is a matrix that is filled with $F$ vectors of $x_t$ where $t$ is the timestep in *obs*, and $W_t \in \mathbb{R}^{d \times F}$ and $B \in \mathbb{R}^{d \times H}$ are to be trained.

Both of the conditions $z_{mix}$ and $_{ar}$ are combined as:

$$\mathbf{c} = \text{concat}([z_{mix}, z_{ar}]) \in \mathbb{R}^{2d \times F} \tag{64}$$

Furthermore, the objective function is set to reduce the error in predicting the data directly (26) instead of the noise (25). The paper shows empirically that predicting the data performs better and hypothesizes that it is due time-series containing quite some noise inherently.

$$\mathcal{L}_{x_\theta} = \mathbb{E}_{k, X_{tar}^0, \epsilon} \left[ ||X_{tar}^0 - x_\theta(X_{tar}^k, k|\mathbf{c})||^2 \right] \tag{65}$$

Shen and Kwok [70] evaluates its model TimeDiff on nine real world datasets: NordPool [102][9], Caiso [104][9], Traffic[76][10], Electricity[75][10], Weather[88][10], Exchange [74][10], ETTh1[5], ETTm1[5], Wind[6]. For each model the historical length $H$ is selected from $\{96, 192, 720, 1440\}$ by testing which works best on the validation set. The paper does not specify which is used in the experiments as shown in Table 11.

| Dataset | H | F | MSE | | | | |
|---|---|---|---|---|---|---|---|
| | | | TimeDiff | TimeGrad | CSDI | SSSD[S4] | D3VAE |
| NorPool | ... | 720 | **0.665** | 1.152 | 1.011 | 0.872 | 0.745 |
| Caiso | ... | 720 | **0.136** | 0.258 | 0.253 | 0.195 | 0.241 |
| Weather | ... | 672 | **0.311** | 0.392 | 0.356 | 0.349 | 0.375 |
| ETTm1 | ... | 192 | **0.336** | 0.874 | 0.529 | 0.464 | 0.362 |
| Wind | ... | 192 | **0.896** | 1.209 | 1.066 | 1.188 | 1.118 |
| Traffic | ... | 168 | **0.564** | 1.745 | N/A | 0.642 | 0.928 |
| Electricity | ... | 168 | **0.193** | 0.736 | N/A | 0.255 | 0.286 |
| ETTh1 | ... | 168 | **0.407** | 0.993 | 0.497 | 0.726 | 0.504 |
| Exchange | ... | 14 | **0.018** | 0.079 | 0.077 | 0.061 | 0.200 |

Table 11: TimeDiff Multivariate Forecasting results [70]

Furthermore, the authors perform an ablation study on the future mixup and AR, resulting in a clear win if both are implemented. They even went as far as implementing these features on CSDI [3.3] and SSSD[S4] [3.7] which also showed that by combining the features resulted in the best performance on ETTh1[5] and ETTm1[5]. CSDI with mixup and AR fell short on performance compared to TimeDiff, while SSSD[S4] achieves comparable results. Besides the better conditioning, the performance of using the data prediction model is also shown across the Caiso[9], Electricity[10], Exchange[10] and Etth1[5] datasets. However, the authors do not perform the experiments of SSSD[S4] with mixup, AR, and a data prediction model. The integration of the data prediction model could have potentially led to superior results. Furthermore, the implementation still struggles with a large amount of input features learning the multivariate dependencies. The paper considers using graph neural networks to capture these dependencies. Overall, TimeDiff has demonstrated significant advancements over prior models in forecasting, marking a substantial progression in long-term prediction accuracy.

### 3.10 TSDiff: Predict, Refine, Synthesize: Self-Guiding Diffusion Models for Probabilistic Time Series Forecasting (2023) [71]

TSDiff is a model that is trained unconditionally and employs the diffusion guidance, explained in section 2.3.2, to condition univariate forecasts during inference. Based on models that use guidance [31, 105], the authors propose a self-guidance mechanism that allows the conditioning during inference without an auxiliary network. This makes it applicable to various downstream tasks. It specifically investigates the usability of task-agnostic unconditional diffusion models for forecasting tasks.

---
[9] DEPTS [103]: https://github.com/weifantt/DEPTS
[10] AutoFormer [25]: https://github.com/thuml/Autoformer

The authors have based the architecture on that of SSSD$^{S4}$ with a modification on how the time-series is represented. Since the model is designed for univariate time-series it utilizes the channel dimension by appending lagged time-series from past seasons resulting in an input $x \in \mathbb{R}^{L \times C}$, replacing the $d$ as explained in section 1.1.1. $L$ is the length of the time-series $H + F$, and additional historical data beyond $L$ can be included in the other $C - 1$ channels. The target series becomes $x_{tar}$ and the observations are $x_{obs}$, the small $x$ denoting the univariate series. It is important to understand that the model will be trained on $X_{obs+tar}$, as the intuition behind the self-guidance is that a model that can generate $obs + tar$ sequences, should also be able to reasonably generate sequences $tar$ while being guided towards $obs$. The authors propose two observation self-guidance approaches, namely the mean square self-guidance and the quantile self-guidance.

**Mean Square Self-Guidance:** Following from Equation 32, the guidance term $p(\mathbf{c}|x_{tar}^k)$ is parameterized and modeled as a multivariate Gaussian distribution

$$p_\theta(\mathbf{c}|x^k) = \mathcal{N}(x_{obs}|f_\theta(x^k, k), \mathbf{I}) \tag{66}$$

where $f_\theta$ is a function that approximates $x^0$ given the diffused time-series $x^k$. Now this function can be expressed using the noise prediction model (16) by rearranging Equation 8 as

$$f_\theta(x_{tar}^k, k) = \frac{x_{tar}^k - \sqrt{1 - \alpha_k}\epsilon_\theta(x_{tar}^k, k)}{\sqrt{\alpha_k}} \tag{67}$$

With this function, the MSE loss is calculated as $\text{MSE}(f_\theta(x_{tar}^k, k), x_{obs})$ and then the gradients are calculated against $x_{tar}^k$. Thus by guiding the denoising steps such that it minimizes the MSE of the observed data in the generated series, it also guides it toward the forecasted data.

**Quantile Self-Guidance:** This guidance goes further to minimize the quantile loss, also known the pinball loss. This loss is based on the CRPS metric, and thus takes all quantiles of the distribution into account, making it more refined than the Mean Square Self-Guidance. So instead of minimizing the MSE, the quantile self-guidance minimizes the quantile loss.

$$\mathcal{L}_{quantile} = \max\{\kappa \cdot (x_{obs} - f_\theta(x^k, k)), (\kappa - 1) \cdot (x_{obs} - f_\theta(x^k, k))\} \tag{68}$$

where $\kappa \in (0, 1)$ specifies the quantile level. In practice Kollovieh *et al.* [71] used multiple evenly spaced quantile levels, based on the number of forecasts. Because the uncertainty is estimated by forecasting multiple times.

Kollovieh *et al.* [71] perform forecasting experiments on eight univariate time series datasets available through GluonTS[2]: Solar[74], Electricity [75], Traffic [76], Exchange [74], M4 [106], UberTLC[77][11], KDD-Cup[107], and Wikipedia[78]. The evaluation method used is CRPS (3) and the distributions were gathered by forecasting 100 times.

The model is designed and evaluated for univariate time-series opposed to multivariate time-series as all of the other models. The authors mention that the model can be extended with a Transformer [21] model to support multivariate time-series by operating across the feature dimensions after the S4 layer. However, the model could encounter limitations when extended to multivariate data because the lagged features of historical data are replaced with the other time-series denoted with $d$, reducing the depth of historical context available. The TSDiff model using quantile self-guidance performs well compared to CSDI [3.3], however, the paper does not compare the results with SSSD$^{S4}$ [3.7] on which the model is based. Instead the authors compare to TSDiff-Cond, a conditional model closely related to SSSD$^{S4}$ and which is more tailored for univariate time-series. TSDiff-Cond has adjustments in the loss computation and the architectural placing of the S4 layers. Furthermore, the authors present research of using unconditional diffusion models for refining existing predictions of other forecasting models and training models on synthetic data, thereby extending the scope beyond just forecasting. To conclude, the proposed models are better than CSDI in the univariate setting and the self-guidance is a novel approach. Also, it would be interesting to see a multivariate comparison with more recent forecasting models and the TSDiff variants.

---

[11] FiveThirtyEight: https://github.com/fivethirtyeight/uber-tlc-foil-response

| Dataset | H | F | CRPS | | | |
|---|---|---|---|---|---|---|
| | | | CSDI | TSDiff-Cond | TSDiff-MS | TSDiff-Q |
| Solar | 336 | 24 | $0.352 \pm 0.005$ | $\mathbf{0.338 \pm 0.014}$ | $0.391 \pm 0.003$ | $0.358 \pm 0.020$ |
| Electricity | 336 | 24 | $0.054 \pm 0.000$ | $0.050 \pm 0.002$ | $0.062 \pm 0.001$ | $\mathbf{0.049 \pm 0.000}$ |
| Traffic | 336 | 24 | $0.159 \pm 0.002$ | $\mathbf{0.094 \pm 0.003}$ | $0.116 \pm 0.001$ | $0.098 \pm 0.002$ |
| Exchange | 360 | 30 | $0.033 \pm 0.014$ | $0.013 \pm 0.002$ | $0.018 \pm 0.003$ | $\mathbf{0.011 \pm 0.001}$ |
| M4 | 312 | 48 | $0.040 \pm 0.003$ | $0.039 \pm 0.006$ | $0.045 \pm 0.000$ | $\mathbf{0.036 \pm 0.001}$ |
| UberTLC | 336 | 24 | $0.206 \pm 0.002$ | $0.172 \pm 0.008$ | $0.183 \pm 0.007$ | $\mathbf{0.172 \pm 0.005}$ |
| KDDCup | 312 | 48 | $0.318 \pm 0.002$ | $0.754 \pm 0.007$ | $0.325 \pm 0.028$ | $\mathbf{0.311 \pm 0.026}$ |
| Wikipedia | 360 | 30 | $0.289 \pm 0.017$ | $\mathbf{0.218 \pm 0.010}$ | $0.257 \pm 0.001$ | $0.221 \pm 0.001$ |

Table 12: TSDiff Univariate Forecasting results [71]

# 4 Discussion

The surveyed papers go into multiple interesting directions tackling time-series forecasting through the use of diffusion denoising processes. Most of the models provide probabilistic forecasting, thereby giving more insights in the uncertainty of the predictions. Notably, the D$^3$VAE [67] and DiffLoad [17] papers advance this field by dissecting both epistemic and aleatoric uncertainties within predictions. DiffLoad does so by performing the diffusion process on a latent space variable of the historical data [17]. Extending on the encoding of historical data, a progression is observed in the methods, evolving from RNNs to Transformers, and ultimately to S4 layers, with the latter demonstrating superior performance in foundational models like [95]. Specifically The SSSD$^{S4}$ [69] and TSDiff [71] models experiment with this type of layer. Additionally, TSDiff reveals that unconditional forecasting with guidance can be as good as conditional forecasting and can bring more benefits for other time-series related tasks. With respect to the type of prediction model used in the reverse process, TimeDiff [70] empirically shows the superiority of data prediction models over noise prediction models for forecasting. This is only a recent revelation and not mentioned in any of the other Furthermore, TimeDiff makes a good comparison with other state-of-the-art transformer forecasting models and shows the superiority of diffusion models by introducing additional indicative bias in the conditioning module. Furthermore, throughout the papers, comparative evaluations utilizing the metrics CRPS$_{\text{sum}}$ (5) or MSE (1) on diverse datasets like Electricity [75], Traffic [76], or those from Gluon-TS[2], have made the comparisons more straightforward and accessible. Lastly, the only implementations that make use of the more generalized diffusion process with the use of SDEs are ScoreGrad [62] and CSPD [66]. With ScoreGrad going as far as to also implement the ODE method enhancing prediction speeds by up to 4.9 times. Overall, this literature survey provides an in-depth overview of 11 seminal papers in the realm of diffusion-based time-series forecasting models. It serves as an invaluable starting point for future researchers delving into this domain, offering a comprehensive understanding of the current state-of-the-art methodologies and their evolutionary trajectories.

# 5 Future Works

Future research directions should include the implementation of diffusion models using the ODE method to enhance prediction speeds without compromising accuracy. The adoption of encoder-decoder frameworks for latent space diffusion, moving beyond mere historical data encoding, is also recommended as this has show to improve prediction qualities [35]. Continued investigation into the S4 layers [91] is encouraged, given their promising results in surpassing Transformers in foundation models [95]. Overall, the integrating of various approaches, such as TimeDiff [70], DiffLoad [17], SSSD$^{S4}$ [69], TDSTF [68], and DSPD-GP [66], would be beneficial. This integration could focus on combining extra conditioning information 3.9, latent space diffusion 3.8, the use of SSMs 3.7, efficient data representation 3.6, and modeling time-series as continuous functions with time-dependent noise 3.4. Attention should also be given to long-term multivariate time-series forecasting, decomposing epistemic and aleatoric uncertainty, and evaluating models with both data and noise prediction models to establish when each approach is more advantageous.